\documentclass{article}

 \usepackage[preprint]{neurips_2026}

\usepackage[utf8]{inputenc} % allow utf-8 input
\usepackage[T1]{fontenc}    % use 8-bit T1 fonts
\usepackage{hyperref}       % hyperlinks
\usepackage{url}            % simple URL typesetting
\usepackage{booktabs}       % professional-quality tables
\usepackage{amsfonts}       % blackboard math symbols
\usepackage{amsmath}
\usepackage{nicefrac}       % compact symbols for 1/2, etc.
\usepackage{microtype}      % microtypography
\usepackage{xcolor}         % colors
\usepackage{mathabx}
\usepackage{cleveref}
\usepackage{amssymb}
\usepackage{float}

\usepackage{pgfplots}
\pgfplotsset{compat=1.18}
\usepackage{tikz}
\usetikzlibrary{external}
\usepackage{multirow}
\usepackage{subcaption}
\usepackage{algorithm}
\usepackage{algpseudocode}

\usepackage{colortbl}
\usepackage{xcolor}
\usepackage{booktabs}
\usepackage{array}
\usepackage{xspace}

\definecolor{heatmap}{RGB}{120, 195, 130}     % soft green
\definecolor{groupbg}{RGB}{238, 238, 244}     % subtle gray for group headers

\newcommand{\hm}[1]{\cellcolor{heatmap!#1!white}#1}
\newcommand{\hmb}[1]{\cellcolor{heatmap!#1!white}\textbf{#1}}
\newcommand{\hmz}{\textcolor{black!30}{0}}

\definecolor{arrowup}{RGB}{15, 85, 220}
\definecolor{arrowdn}{RGB}{200, 60, 60}
\newcommand{\au}{\,\textcolor{arrowup}{\scriptsize$\uparrow$}}
\newcommand{\ad}{\,\textcolor{arrowdn}{\scriptsize$\downarrow$}}

\newcommand{\hmu}[1]{\cellcolor{heatmap!#1!white}#1\au}
\newcommand{\hmd}[1]{\cellcolor{heatmap!#1!white}#1\ad}
\newcommand{\hmub}[1]{\cellcolor{heatmap!#1!white}\textbf{#1}\au}
\newcommand{\hmdb}[1]{\cellcolor{heatmap!#1!white}\textbf{#1}\ad}
\newcommand{\hms}[1]{\cellcolor{heatmap!#1!white}#1\textsuperscript{*}}

\newcommand{\hmzd}{\textcolor{black!30}{0}\ad}

\definecolor{opuscolor}{RGB}{91, 122, 184}
\definecolor{gptcolor}{RGB}{217, 130, 70}
\definecolor{geminicolor}{RGB}{99, 165, 105}
\definecolor{highcolor}{RGB}{61, 90, 137}
\definecolor{nonecolor}{RGB}{180, 180, 180}

\definecolor{targetc}{RGB}{95,165,100}
\definecolor{decoyc}{RGB}{225,115,90}
\definecolor{predc}{RGB}{95,130,200}
\definecolor{fillerc}{RGB}{230,230,230}

\definecolor{c_locate}{RGB}{44,160,44}      % green
\definecolor{c_longest}{RGB}{31,119,180}    % blue
\definecolor{c_scatt}{RGB}{214,39,40}       % red
\definecolor{c_clust}{RGB}{255,127,14}      % orange
\definecolor{c_multi}{RGB}{148,103,189}     % purple
\definecolor{c_uspar}{RGB}{23,140,140}      % teal
\definecolor{c_udist}{RGB}{227,119,194}     % pink

\newcommand{\advlongbench}{\textsc{PredicateLongBench}\xspace}

\title{Understanding Axes of Difficulty For Long Context Tasks Via \advlongbench}

\author{%
  Siddhartha Jain \\
  NVIDIA \\
  \texttt{siddjain@nvidia.com} \\
  \And
  Ameya Velingker \\
  NVIDIA \\
  \texttt{avelingker@nvidia.com} \\
}

\begin{document}

\maketitle

\begin{abstract}
Large language models (LLMs) have demonstrated rapidly improving long-context capabilities, prompting a wave of benchmarks designed to evaluate them. However, existing long-context evaluations—from Needle-in-a-Haystack (NIAH) tests to more recent multi-hop reasoning and summarization tasks—predominantly measure average-case performance, and many are either saturated or lack robustness. Notably absent is a systematic way to probe how models perform as we scale up the difficulty of tasks along various axes. We address this gap by proposing \advlongbench, a benchmark that stress-tests long-context reasoning by asking models to identify the longest contiguous subsequence of words in a long input that satisfies given predicates/constraints (e.g., lexicographic ordering), drawn from a broader predicate class. The central innovation of our benchmark is the identification and systematic exploration of multiple different axes of difficulty which test multiple aspects of long context understanding. We provide two complementary generation pipelines — a fully synthetic setup using random word-like strings, and a real-world setup that samples words from natural documents while preserving their distributional properties. We find that frontier models struggle to perform well as we scale up the difficulty of tasks along our axes, demonstrating the utility of our benchmark in understanding the limitations of current long-context capabilities. Furthermore, the tasks in \advlongbench, though challenging, are conceptually simple and do not require LLM-based generations or judges.
\end{abstract}

\section{Introduction}
Large language models (LLMs) have shown increased long-context performance in recent years. As more and more powerful models capable of processing even longer context lengths emerge, there is an increasing need to have robust benchmarks for evaluating long-context capabilities of a model.

One of the earliest markers of long-context performance has been the popular Needle-in-a-Haystack (NIAH) test, which evaluates a model's ability to find a ``needle'' (tiny piece of localized information) hidden inside a ``haystack'' (a large window of context)~\citep{Kamradt23,MohtashamiJ23}. A number of variants of NIAH have been devised to make the task more difficult, e.g., embedding multiple needles, adding distractors to the context, etc. However, NIAH tasks nevertheless have limitations and are generally restricted to locating small bits of highly localized information. The need to measure a broader range of long-context capabilities has spurred the development of other long-context evaluations that incorporate additional task diversity in the form of question answering (QA), multi-hop reasoning, summarization, etc.

There are a number of axes along which researchers have sought to increase the difficulty of long-context evaluations. The most natural one is \emph{context length}, as increasing the underlying context length of tasks requires a model to find or reason on information contained in a much larger sea of text. To that effect, many popular long-context benchmarks, e.g., RULER~\citep{HsiehSKARJZG24}, HELMET~\citep{YenGHDFIWC25}, are configurable and adaptable to any desired context length, thus allowing one to test the degradation in performance over an increasing list of context lengths.

Another axis is the use of \emph{distractors}, which are surrounding blocks of text designed to hide the key information in the context window. For example, early NIAH tests, such as those in RULER, have used blocks of irrelevant content (e.g., filler phrases like ``The grass is green,'' sentences from documents or essays, etc.) to obfuscate the needles (key-value pairs). Newer NIAH tasks, such as those in NoLiMa~\citep{ModarressiDDBRYS25}, use more sophisticated distractors, designed to be semantically similar to the relevant text. Distractors are also found in other types of tasks such as QA, where topically similar paragraphs can be dispersed throught the context.

\emph{Computation} is yet another axis along which difficulty may be scaled. A long-context task may be made more difficult by requiring it to involve additional computational steps. For instance, many QA tasks are explicitly designed to be multi-hop tasks, thus requiring a model to obtain successive pieces of information from different parts of the context, with each successive query guided by the result of the previous retrievals. Other long-context tasks that incorporate computation include cross-document QA, document summarization, and long codebase understanding.

A related axis to computation is \emph{non-locality}, which refers to the concept that a given task requires a model to examine the entire context window to be able to arrive at the correct answer. NIAH tasks are notoriously local, i.e., the key-value pair is contained inside a small local window of the context, and once the requisite key is located, there is no need for the model to see the rest of the context. On the other hand, some tasks are highly non-local, such as summarization or frequency counting tasks (e.g., CWE/FWE in RULER). We refine this notion in the form of \textit{quantifier complexity} later in the paper.

Many long-context evaluation benchmarks scale difficulty along some axes but not others. Moreover, scaling difficulty often comes from making tasks much more complex, e.g., requiring the use of LLMs to generate the context/tasks or judge responses. Motivated by this state of affairs, we present \advlongbench, a new family of tasks that retains simplicity while combining different approaches to scale difficulty across multiple axes.

% A large number long-context benchmarks center on tasks that are either saturated or lack robustness. While measuring "average case" or real-world long-context performance is a useful metric, it is also desirable to study how models perform on \emph{adversarial} long-context tasks. Surprisingly, there is a dearth of long-context benchmarks that measure \emph{adversarial} long-context performance.

Our main contributions are as follows: (1) We present \advlongbench, a family of tasks that are algorithmically simple yet pose a challenge for frontier LLMs, (2) We systematically define and explore multiple axes for increasing task difficulty and present variants of simple retrieval tasks that scale the difficulty across these axes, (3) We devise predicates for constraint satisfaction that allow us to scale the difficulty of simple retrieval tasks along these axes, (4) We conduct experiments on the structure of the context (specifically distractors) and show that the structure has a large impact on frontier model peformance on these tasks, (5) We explore increasing difficulty along a new axis, namely, search space, which can be scaled up even while keeping context token budgets fixed.

\section{Related Works}
Several long-context benchmarks have been proposed which consist of a diverse suite of tasks. We highlight such works below.

\paragraph{Advanced NIAH tasks.}
Several benchmarks build on the vanilla NIAH paradigm, adding additional complexities to make tasks more challenging. For instance, NoLiMa~\citep{ModarressiDDBRYS25} constructs NIAH tasks with carefully-designed sets of needles designed to minimize lexical overlap with the haystack. Other recent NIAH benchmarks explore ideas around parallelism (NeedleThreading, \citep{RobertsHA24}), multilingual retrieval (MLNeedle, \citep{HengleBDC25}), in-context features (DENIAHL, \citep{DaiPYBM24}), code repositories (RepoQA, \citep{LiuTDWDWYZ24}), etc.

\paragraph{Multi-Round Co-reference Resolution (MRCR).} The \emph{Multi-Round Co-reference Resolution (MRCR)} task, introduced in \citet{Gemini-1.5-24}, presents as context a conversation between a user and a model (in which a user requests a model to write essays, poems, etc. on different topics), then requiring a model to reproduce the output from the conversation according to a given request. This may be viewed as an advanced NIAH task in which the relevant conversation is the ``needle,'' which is prompted by a key (the request). MRCR forms the basis for popular benchmarks such as, e.g., Michelangelo~\citep{VodrahalliOTXJSHDK+24}, OpenAI-MRCR~\citep{openai2025mrcr}, etc.

\paragraph{Graph-Based Benchmarks.} A more recent direction in long-context evaluations has been the exploration of tasks on graphs. Graphs offer a natural route toward constructing multi-hop tasks. For instance, in the OpenAI-Graphwalks~\citep{openai2025graphwalks} benchmark, a model is provided with a list of edges of a large graph, and the tasks involve graph traversal (e.g., breadth-first search). The benchmark adds complexity beyond that of standard multi-hop tasks because a single sequential graph traversal is not sufficient to solve the tasks.

\paragraph{Real-World Benchmarks.} While many of the aforementioned benchmarks involve \emph{synthetic} tasks, there has been recent interest in devising \emph{real-world} tasks. Such tasks generally fall into a number of categories such as Question Answering (QA), summarization, in-context learning, document retrieval and reranking, retrieval-augmented generation (RAG), code tasks, etc. Benchmarks that focus on real-world long-context tasks include HELMET~\citep{YenGHDFIWC25}, LongBench~\citep{BaiLZLTHDLZHDTL24}, LongBench v2~\citep{BaiTZPWLCXHDTL25}, InfinityBench~\citep{ZhangCHXCHHTWLS24}, Long Code Arena~\citep{BogomolovEGGSTGKDIB24}, NoCha~\citep{KarpinskaTLGI24}, etc.

\paragraph{Other Metrics.}
In addition to long-context benchmarks, many have also studied standard metrics such as perplexity as a proxy for long-context performance. While some have used a low perplexity score as an indication of successful context window extensions of a model, some studies point out a large gap between perplexity and performance on actual long-context tasks~\citep{HuHTZF24,HsiehSKARJZG24}. Some works have sought to rectify the gap by making suitable modifications to perplexity. For instance, \citet{FangWLZJGDW25} introduce LongPPL, a weighted average over perplexity of tokens that assigns higher weight to selected key tokens (that are dependent on longer context), as well as LongCE loss, a re-weighting strategy for fine-tuning.

% \subsection{Comparison to Distractors in Other Tasks}
% We point out that there are various other long-context evaluations whose tasks involve the addition of distractors. We outline these and also describe how our decoy setup in \advlongbench is conceptually different.

% A number of tasks add distractors to NIAH tasks. For instance, RULER uses distractor key/value pairs in its NIAH tasks. However, these are not designed to be adversarially similar to the correct key/value pair. NoLiMa additionally tries to use topically similar distractors.

\section{\advlongbench Benchmark}

\subsection{Task Setup}
The general setup is to provide a long context window that consists of a space separated list of words. A typical task in our benchmark asks a model to identify the longest contiguous subsequence of words subject to constraints, chosen from some predicate class.

More precisely, suppose we have a predicate class $\mathcal{H} \subseteq \{(a_n) \mapsto \{0, 1\} \}$. A task, defined by a list of predicates $f_1, f_2, \dots, f_j \in \mathcal{H}$, is to identify the largest $k$ such that there is a contiguous subsequence of words $w = (w_1, w_2, \dots, w_k)$ subject to $f_i(w) = 1$ for $i = 1, 2, \dots, j$.

\subsection{Predicates}
The tasks as described earlier are quite general. The specific tasks are defined by the choice of predicates. We discuss specific predicates here.

\subsubsection{Unary predicates}
These are predicates that depend only on a single word, i.e., $f(w_1, w_2, \dots, w_k) := f(w_i)$ for a single $i$. Tasks with only unary predicates are the simplest type of task, as the predicates operate independently on individual words in the sequence.

\paragraph{Prefix/Suffix/String Match unary predicate tasks.} In this work, we consider unary predicates that are related to string containment. In particular, we consider the functions $\mathrm{pre}_{s, i}, \mathrm{suff}_{s, i}, \mathrm{cont}_{s,i}$:
\begin{itemize}
    \item $\mathrm{pre}_{s, i}(w) = 1$ if and only if string $s$ is a prefix of $w_i$.
    \item $\mathrm{suff}_{s, i}(w) = 1$ if and only if string $s$ is a suffix of $w_i$.
    \item $\mathrm{cont}_{s, i}(w) = 1$ if and only if string $s$ is contained in $w_i$.
\end{itemize}

\subsubsection{Binary predicates}
In this case, each predicate depends on up to two words of the sequence, i.e., $f(w) := f(w_{i_1}, w_{i_2})$ for two indices $i_1, i_2$. Such predicates allow for more challenging tasks, as they model interactions between two different words in the sequence.

\paragraph{Lexicographic predicates.} In this work, we consider tasks which involve outputting sequences of words that are in lexicographic order. The \emph{lexicographic predicate} $f$ satsifies $f(w_1, w_2, \dots, w_k) = 1$ if and only if $w_1 \preccurlyeq w_2 \preccurlyeq \cdots \preccurlyeq w_k$, where $x \preccurlyeq y$ indicates that $y$ occurs no earlier than $x$ in the standard lexicographic ordering. $f$ can be decomposed into binary predicates as follows: $f(w_1, w_2, \dots, w_k) = 1$ if and only if 
\[
 \mathrm{lex}_{i}(w) = 1 \quad\text{for all } i=1,2,\dots, k-1,
\]
where $\mathrm{lex}_{i}(w)$ is 1 if and only if $w_i \preccurlyeq w_{i+1}$.
Intuitively, a sequence is in lexicographic order if and only if pairs of adjacent words are in order.

\subsection{Axes of difficulty}

As mentioned in the Introduction, a major contribution our work is defining and systematically exploring various axes of difficulty for the tasks. We outline these axes in the context of our task setup below.

\subsubsection{Computation}

A key aspect of our tasks is that retrieval requires some level of computation as one needs to check whether the retrieved sequence satisfies the predicate. This opens up the possibility of scaling difficulty by increasing the level of computation required for that retrieval. In particular, by increasing the \textit{arity} of the predicate, we can increase its computational complexity. In our case, we test arity 1 (unary) and arity 2 (binary) predicates.

\subsubsection{Adversarial Decoys}
Here the idea is to ensure that the context contains continguous subsequences that almost match the desired predicate except for a tiny deviation. We experiment with two decoy variants.

\paragraph{Near-Miss Decoys.} 
For unary predicates, we use \textit{near-miss} decoys where for $\alpha = 0.05$ fraction of words, we modify them to satisfy a predicate at random. We do this for all predicates (thus total fraction of words modified is $\alpha\cdot k = 0.25$ in our case) while making sure that no two words are modified to satisfy the same predicate and there still remains a unique sequence that satisfies all the predicate criteria.

For lexicographic tasks we use \textit{near-sorted} sequences. In addition to the correct sequence that must be output, we ensure that the context contains a number of such sequences (8 in our case) of the same length, i.e., sequences that are lexicographically valid and of length $k$ except they have a single swapped interior pair. The window's flanks are chosen so that the surrounding context cannot extend the run on either side, so the decoy has the same length as the target and visually resembles it; only a careful pairwise check across the entire window reveals the inversion. The exact construction differs slightly between the synthetic and real-document settings, and is given in Appendix~\ref{app:tasks}.

\paragraph{Multi-List Decoys.} A complementary decoy variant tests the model's ability to respect a structural constraint rather than just local sequence properties. We partition the word list into $M$ contiguous numbered sublists and plant a length-$k$ sorted run across every adjacent sublist boundary, so that the suffix of one sublist together with the prefix of the next forms a valid lex-sorted sequence. The model is instructed to return a run contained \emph{within} a single sublist, so each cross-boundary run is lexicographically valid as a raw subsequence yet violates the within-sublist constraint of the task. The unique target run of length $k$ is planted entirely within one randomly chosen sublist. Construction details are given in Appendix~\ref{app:tasks}.

\subsubsection{Search Space Size}

We consider search space size from the perspective of how much \textit{search} is needed by an exact algorithm to accomplish the task. As our tasks involve retrieval of a sequence of words from a list of words, in general if the list of words increases -- even if the total token count remains the same -- that would increase the search space (and thus running time) for an exact algorithm. We thus experiment with a variant of the lexicographic sequence retrieval task where we keep the token count fixed but increase the search space by using shorter words that use fewer tokens.

\subsubsection{Quantifier complexity}

Our baseline unary and binary predicate tasks ask about locating a sequence satisfying some constraints. In formal logic terms, the query involves an \textit{existential}($\exists$) quantifier. We can however construct tasks that involve a \textit{universal}($\forall$) quantifier which is typically a harder query to answer. In our case, we construct two such variants. For the unary predicate task, instead of asking for any sequence that satisfies the given set of predicates, we ask for \textit{all} such sequences. Even though in our data, there is only one such sequence, this forces the model to reason about the entire list of words it has been given and make sure that for all other sequences, none of them satisfy the predicates. For the lexicographic sequence retrieval task, instead of asking for a lexicographic sequence of a particular length, we ask for a maximum length such sequence. This again forces the model to evaluate and reject every possible sequence that is invalid.

\subsubsection{Context structure hardness}

Unstructured spaces are much harder to navigate. We directly test this axis by constructing more structured versions of the context. Specifically for the near-sorted lexicographic decoy tasks, we create task variants where the decoys are \textit{clustered} together along with the target sequence as opposed to being randomly scattered throughout the context.

\subsection{Generation}

We now have the pieces in place to describe the set of tasks. As mentioned before, the task setup is retreiving a sequence subject to given predicates. The axes of difficulty for constructing the task variants can modify the query or the context on top.

\subsubsection{Source of list of words}

\paragraph{Fully synthetic} In this setup we sample words as fixed-length random strings over the lowercase English alphabet (length $8$ for
lexicographic tasks, $12$ for unary tasks). Words are drawn uniformly without replacement by sampling integers from $\{0, \dots, 26^L - 1\}$ and decoding each in base $26$, yielding a uniform distribution over word space. Target sequences and decoys are then inserted at chosen positions in a shuffled list; any unintended lex run of length $\geq k$ that arises by chance is destroyed via a swap-based fixup, so each instance contains exactly one valid solution. We target approximately 128K tokens in the generated list as per Qwen's tokenizer and construct 100 examples per variant. The detailed algorithm is given in Appendix~\ref{app:tasks}.

\paragraph{Real-World.} Concretely, we draw a set of documents from LongBench~v2 \citep{BaiTZPWLCXHDTL25} and apply the filtering rules of
Appendix~\ref{app:longbench} to discard documents whose tokenized form contains artifacts that would trivialize the task --- a long lex-sorted run already present in the source, a repetitive vocabulary, HTML markup, and the like. This leaves us with 120 documents from which we subsample 100. Each surviving document is tokenized into a flat list of words, and this list is used directly as the base context. Because the base list is the document's own word stream, its natural lexical distribution is preserved by construction. The number of tokens in context can go upto 365K (as per Qwen's tokenizer). As the open-source models we test do not have as long a context window, we subselect for documents with <170K tokens. This gives us 93 documents and thus we evaluate on 93 examples per variant for open-source models for LongBench sourced tasks.

\subsection{Set of Tasks}

\begin{figure}[H]
\centering
    \includegraphics[width=0.8\textwidth]{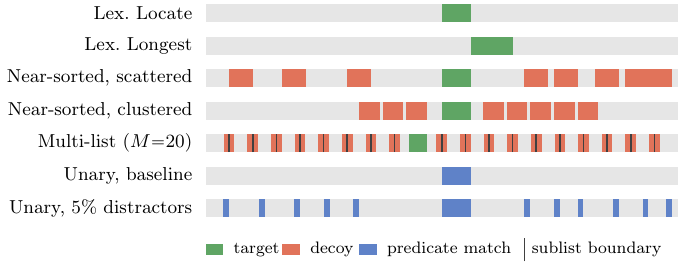}
\caption{Schematic of \advlongbench task variants. Each strip represents a
long-context word list (light grey) for one task
instance. Highlighted regions mark the unique target run
(\textcolor{targetc}{green}); decoys
(\textcolor{decoyc}{red}); and predicate-matching positions for unary tasks
(\textcolor{predc}{blue}). In the multi-list row, vertical bars indicate
the $M = 20$ sublist boundaries, and the \textcolor{decoyc}{red} blocks
straddling each boundary are cross-boundary decoy runs that are
lexicographically valid as raw subsequences but lie on the wrong side of
the within-sublist constraint. In the bottom row, scattered
\textcolor{predc}{blue} cells are isolated distractor predicate matches
that do not form a valid contiguous window. The same schematic applies to
both the synthetic and LongBench-based instantiations of each task; only
the underlying word source differs. Block sizes and densities are
exaggerated relative to the true list length for visibility.}
\label{fig:tasks}
\end{figure}

We present the full list of tasks here (see \Cref{tab:tasklist}). A visualization of the task variants is also given in \Cref{fig:tasks} :

\begin{table}[t]
\centering
\small
\renewcommand{\arraystretch}{1.2}
\setlength{\tabcolsep}{6pt}

\caption{Tasks in \advlongbench. Each row is a single variant; \textbf{Axis} names the difficulty axis the variant perturbs relative to
its family's baseline (italicized). \textbf{Synth.}\ and \textbf{Real} indicate whether the variant is generated from synthetic words
(Alg.~\ref{alg:sample-words}) or from filtered LongBench~v2 documents (\S\ref{app:longbench}); \checkmark\ = available, --- = not run.$\exists$ is the existential quantifier and $\forall$ is the universal quantifier. $k=5$ for all synthetic tasks and equal to the longest lexicographic run for LongBench tasks.}.
\label{tab:tasklist}

\begin{tabular}{@{}llcc@{}}
\toprule
\textsf{\bfseries Variant} & \textsf{\bfseries Axis varied} 
& \textsf{\bfseries Synth.} & \textsf{\bfseries Real} \\
\midrule

\rowcolor{groupbg}
\multicolumn{4}{@{}l}{\textit{\textbf{Unary-predicate family} \,---\, baseline: $\exists$-query, no distractors}} \\
$\exists$-query (any seq.)                  & \textit{baseline}             & \checkmark & --- \\
$\forall$-query (all seqs.)                 & Quantifier                    & \checkmark & --- \\
$\forall$-query, 5\% per-position distractors ($\alpha=0.05$)  & Decoys ($+$ quantifier)       & \checkmark & --- \\

\rowcolor{groupbg}
\multicolumn{4}{@{}l}{\textit{\textbf{Lexicographic family} \,---\, baseline: locate-$k$, $\exists$-query, no decoys}} \\
Locate-$k$                                  & \textit{baseline}             & \checkmark & \checkmark \\
$+$ near-sorted decoys (8 of them)          & Decoys                        & \checkmark & \checkmark \\
$+$ near-sorted decoys (clustered) & Context structure             & \checkmark & \checkmark \\
$+$ multi-list (20 sublists) decoys         & Decoys (structural)           & \checkmark & --- \\
Longest                           & Quantifier                    & \checkmark & \checkmark \\
Locate-$k$, larger search space             & Search space                  & \checkmark & --- \\

\bottomrule
\end{tabular}
\end{table}

In particular, our tasks fall into two main categories: unary predicates and binary predicates, the latter consisting solely of lexicographic tasks. For both tasks, we have a baseline variant as well variants where we move along various axes of difficulty. We use $k=5$ for all task variants except for LongBench tasks where $k$ is set to the length of the longest lexicographic sequence which can vary from 6 to 10. A complete description of task specifications is found in Appendix~\ref{app:tasks}.

\section{Results}

\begin{table}[t]
\centering
\small
\renewcommand{\arraystretch}{1.25}
\setlength{\tabcolsep}{4pt}

\caption{Main results on \advlongbench, organized by axis of difficulty.
The top section has the natural baselines; subsequent sections perturb it
along a single or pair of axes. Cells are shaded by accuracy
(white = 0\%, green = 100\%); \textbf{bold} marks the best entry per row.
Arrows compare each cell to its closest reference in the top section.
\textcolor{arrowup}{$\uparrow$} = up vs.\ reference,
\textcolor{arrowdn}{$\downarrow$} = down. Closed reasoning models use
\emph{high}, low (1024-token reasoning budget), and \emph{none} effort. The context length of synthetic tasks was approximately 128K and upto 365K for the LongBench tasks as per the Qwen tokenizer. The
max output tokens is set to 16K. Each model was run on 100 examples per
variant except for open-source models on LongBench sourced tasks where
they were run on 93 examples per variant. \\
\textsuperscript{*}{\small \bf On the unary $\exists$ baseline, Opus's
non-matching outputs are entirely refusals rather than incorrect answers,
so the reported accuracy understates its capability and the apparent gain
on the $\forall$-row in (2) is an artifact of this.}}
\label{tab:main_results}

\resizebox{\textwidth}{!}{%
\begin{tabular}{@{}lcccccccccc@{}}
\toprule
& \multicolumn{2}{c}{\textsf{\bfseries Opus 4.6}}
& \multicolumn{2}{c}{\textsf{\bfseries GPT-5.4}}
& \multicolumn{2}{c}{\textsf{\bfseries Gemini 3.1}}
& \textsf{\small \bfseries MiniMax}
& \textsf{\small \bfseries GLM}
& \textsf{\small \bfseries Qwen 3.5} \\
\cmidrule(lr){2-3} \cmidrule(lr){4-5} \cmidrule(lr){6-7} \cmidrule(lr){8-10}
\textsf{\small Variant}
& \textsf{\small high} & \textsf{\small none}
& \textsf{\small high} & \textsf{\small none}
& \textsf{\small high} & \textsf{\small low}
& \textsf{2.7} & \textsf{5.1} & \textsf{397B} \\
\midrule

\rowcolor{groupbg}
\multicolumn{10}{@{}l}{\textit{\textbf{(1) Baselines} \,---\, computation axis (predicate arity $\times$ word source)}} \\
Unary, $\exists$-query (any seq.)              & \hms{87} & \hms{53}& \hm{95}  & \hmb{96}& \hm{93} & \hm{91} & \hm{50} & \hm{76} & \hm{45} \\
Lex.\ locate, synthetic                        & \hmb{97} & \hm{25} & \hm{62}  & \hm{2}  & \hm{43} & \hm{45} & \hm{8}  & \hm{8}  & \hmz    \\
Lex.\ locate, LongBench                        & \hmb{61} & \hm{10} & \hm{36}  & \hm{1}  & \hm{6}  & \hm{10} & \hmz    & \hm{3}  & \hmz    \\

\rowcolor{groupbg}
\multicolumn{10}{@{}l}{\textit{\textbf{(2) Quantifier complexity} \,---\, $\exists$ $\to$ $\forall$ over the same task}} \\
Unary, $\forall$-query (all seqs.)             & \hmub{99}& \hmu{90}& \hmd{93} & \hmd{93}& \hmd{59}& \hmd{20}& \hmd{30}& \hmd{73}& \hmd{43} \\
Lex.\ longest, synthetic             & \hmdb{23}& \hmd{5} & \hmd{12} & \hmzd   & \hmd{2} & \hmd{15}& \hmd{2} & \hmzd   & \hmz     \\
Lex.\ longest, LongBench             & \hmdb{20}& \hmd{5} & \hmd{7}  & \hmzd   & \hmzd   & \hm{10} & \hmz    & \hmzd   & \hmz     \\

\rowcolor{groupbg}
\multicolumn{10}{@{}l}{\textit{\textbf{(3) Adversarial decoys} \,---\, decoys / distractors planted, scattered across context}} \\
Unary $\forall$ + 5\% per-position distractors ($\alpha=0.5)$ & \hmdb{7} & \hmzd   & \hmd{5}  & \hmzd   & \hmd{1} & \hmd{5} & \hmzd   & \hmd{2} & \hmzd    \\
Lex.\ locate + near-sorted, synthetic          & \hmd{1}  & \hmzd   & \hmzd    & \hmzd   & \hmdb{2}& \hmzd   & \hmzd   & \hmzd   & \hmz     \\
Lex.\ locate + multi-list (20), synthetic      & \hmd{1}  & \hmzd   & \hmd{2}  & \hmzd   & \hmdb{10}&\hmzd   & \hmzd   & \hmzd   & \hmz     \\
Lex.\ locate + near-sorted, LongBench          & \hmdb{49}& \hmzd   & \hm{36}  & \hm{1}  & \hmu{11}& \hmzd   & \hmz    & \hmd{2} & \hmz     \\

\rowcolor{groupbg}
\multicolumn{10}{@{}l}{\textit{\textbf{(4) Context structure} \,---\, same decoys as (3), clustered near the target}} \\
Lex.\ locate + near-sorted (clustered), synthetic    & \hmub{98}& \hmu{30}& \hmu{70}& \hmu{5} & \hmu{64}& \hmd{5} & \hmu{10}& \hmd{2} & \hmz \\
Lex.\ locate + near-sorted (clustered), LongBench    & \hmub{93}& \hmu{35}& \hmu{68}& \hmu{4} & \hmu{56}& \hmu{50}& \hmz    & \hmu{12}& \hmz \\

\bottomrule
\end{tabular}
}
\end{table}

In this section, we provide results for a number of frontier models on our benchmark, \advlongbench (see \Cref{tab:main_results}). As mentioned earlier, for synthetic tasks the context length is 128K tokens as per the Qwen tokenizer and can go upto 365K for LongBench derived tasks. In general the trend we observe is that for several of the baselines, closed-source frontier models and for some, open-source frontier models perform quite well. However as we increase difficulty across our various axes, the performance uniformly crashes.

\subsubsection*{Models evaluated}
We evaluate the following popular models: GPT 5.4 (reasoning: high), GPT 5.4 (reasoning: none), Gemini 3.1 Pro preview (reasoning: high), Gemini 3.1 Pro preview (reasoning: low), Opus 4.6 (reasoning: high), Opus 4.6 (reasoning: none), Minimax 2.7, GLM 5.1, Qwen 3.5 397B. All results are with max output tokens of 16K unless otherwise stated. Details of how they are prompted are deferred to the appendix.

Below, we highlight various insights from the results.

\subsection{Binary predicates are more difficult than unary predicates (Computational axis)}

As shown in \Cref{tab:main_results}, even tasks based on simple binary predicates specifying that consecutive words have to be sorted (lexicographic) are more challenging for models than the task of checking whether consecutive words satisfy particular unary predicates (in our case whether they contain two different particular strings). Our hypothesis is that binary predicates require models to generate representations sensitive to pairwise relationships between consecutive words as opposed to just the individual words independently, and the models do the latter much more consistently than the former.

\subsection{Quantifier hardness}

\subsubsection{Retrieval is harder with universal quantifier query}
We test the models on the task of retrieving \textit{all} sequences satisfying the unary predicates instead of just one. However we still ensure that there is a single such sequence in the context. Thus the answer does not change but the model now has to make sure that for all sequences, none of them satisfy the predicates. The results are in \Cref{tab:main_results}. Despite the true answer not changing between the prompts, most models suffer a big loss in accuracy. The exception is the Opus model, which sees big gains. However as explained in \Cref{tab:main_results} itself, this is artificial, as all of the ``incorrect'' answers for Opus for the baseline variant come from refusal to answer the question.

\subsubsection{Finding the longest sequence is much harder}
Asking the model to retrieve the maximum length lexicographic sequence entails identifying the distribution of lengths of lexicographic sequences and then retrieving the longest one. In particular, it requires maintaining a \textit{memory} that holds the value of the maximum length of a past lexicographic sequence consistently across the entire context, which the task of merely locating a sequence of a given length does not. We test whether the maximum length retrieval task is truly more difficult. As shown in \Cref{tab:main_results}, we find that this task does indeed pose a greater challenge for models.

\subsection{Decoys}

\subsubsection{Near-miss Decoys}
We construct near-miss decoys for both unary and binary (lexicographic) predicates. For the former, we modify 5\% of the words per position in the length 5 sequence so that they satisfy one of the predicates while still ensuring that only a single unique sequence of length 5 satisfies all the predicates in the correct order. For the latter, we insert eight length-5 sequences that would be lexicographically sorted except for a single pair of consecutive words in the middle that is swapped.

\begin{figure}[H]
\centering
\begin{subfigure}{0.49\textwidth}
\centering
\includegraphics[width=\textwidth]{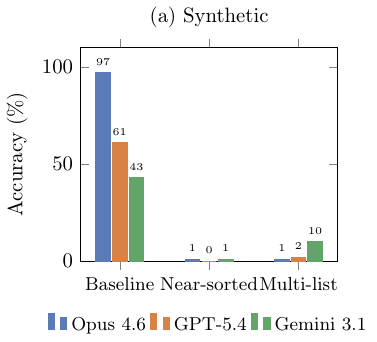}
\end{subfigure}
\hfill
\begin{subfigure}{0.49\textwidth}
\centering
\includegraphics[width=\textwidth]{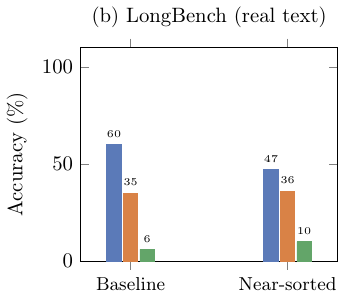}
\end{subfigure}
\caption{Decoys cause dramatic collapse on synthetic data but barely affect LongBench. (a) On synthetic lexicographic locate, near-sorted scattered decoys reduce all three models from $\geq 43\%$ to $\leq 2\%$, and multi-list decoys to $\leq 10\%$. (b) The same scattered near-sorted decoys barely hurt the strongest models on LongBench (and slightly help Gemini), suggesting that natural-text words provide distinguishing signal that synthetic strings lack.}
\label{fig:decoy_collapse}
\end{figure}

The results are in \Cref{fig:decoy_collapse}. For all models, inserting the decoys leads to a drastic drop in performance. For example, for Opus 4.6 with high reasoning, for unary tasks the performance crashes to just 7\% while for GPT 5.4 high, it crashes to 5\%. For the lexicographic task, that state of affairs is even worse; with just 8 near-sorted decoys, no model exceeds 2\%.

\subsubsection{Multi-List decoys}

We construct another decoy variant for the lexicographic sequence retrieval task. In this variant we split our list of words into 20 lists and for every list boundary, we ensure that the suffix of the list and the prefix of the next list form a lexicographically sorted sequence of length 5. However we task the model with retrieving a sequence wholly contained in a single list. The results are in \Cref{tab:main_results}. As with the previous decoy, we see a drastic drop in performance with this decoy as well.  The best performing model is Gemini 3.1 high, which obtains 10\%, while Opus 4.6 and GPT 5.4 are at 1\% and 2\%, respectively, and open-source models are further behind at 0\%.

\subsection{Search space size matters, not just token count}

We additionally assess whether increasing the \textit{search space} for a task while keeping the token count of the context the same leads to performance decline. In the standard synthetic word lists that we construct, we typically have ~26K words in the prompt. For the variant with a larger search space, we construct a synthetic word list with shorter words so that the prompt has ~60K words (while the token length of the prompt is maintained at 128K). We run GPT 5.4 with reasoning effort high and max output token budget of 128K. The performance drops from 92\% (for the 128K output token budget) to just 10\%. This highlights an often underemphasized aspect of long context evaluation --- the search space for the relevant task matters in addition to the raw number of tokens in the prompt.

\subsection{Structured context makes search easier}

\begin{figure}[H]
\centering
\begin{subfigure}{0.49\textwidth}
\centering
\includegraphics[width=\textwidth]{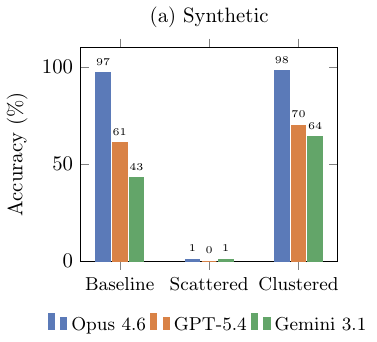}
\end{subfigure}
\hfill
\begin{subfigure}{0.49\textwidth}
\centering
\includegraphics[width=\textwidth]{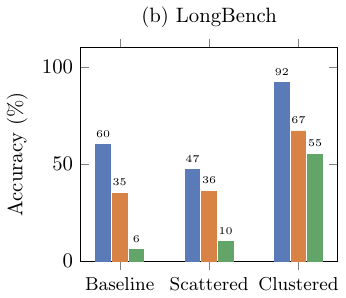}
\end{subfigure}
\caption{Clustering near-sorted decoys near the target sequence \emph{restores} or even \emph{exceeds} baseline accuracy, while the same decoys scattered across the context cause catastrophic collapse. The hardness of near-sorted decoys is therefore a long-context property: when decoys are spatially concentrated, models can localize the relevant region and reason within it.}
\label{fig:clustering}
\end{figure}

If the information in the context is arranged in a more structured way, does that help the model? We set out to test this by \textit{clustering} the near-sorted decoys for the lexicographic sequence retrieval task together in the context along with the ground truth sequence itself. As shown in Figure~\ref{fig:clustering}, this has a dramatic effect on performance, completely restoring and sometimes even exceeding the performance on the vanilla task of retrieving a target length sequence without any adversarial setup. Thus, the hardness of the near-sorted decoys is tied to not just the presence of the decoys but also to the fact that they are scattered across context. We speculate that when they are clustered together, they likely form a distinctive region that lets the models zone in quickly on that region and reason as to which one of the alternatives is the correct sequence. Thus, the adversarialness of the near-sorted decoys is intimately tied to the fact that we are working with long context.

\subsection{More inference time compute improves results}

\begin{figure}[H]
\centering
\includegraphics[width=\textwidth]{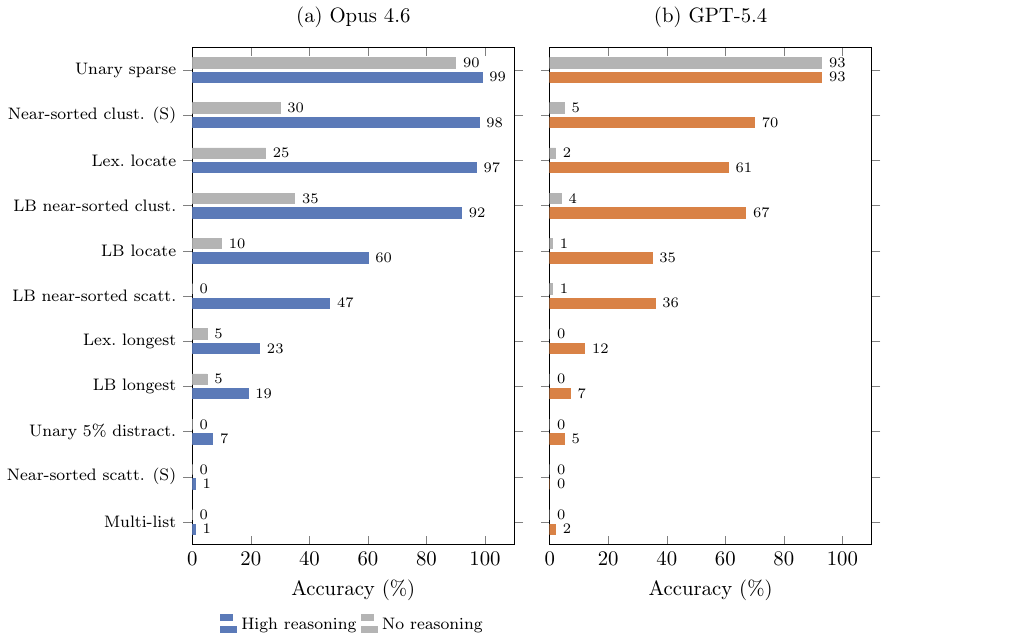}
\caption{Reasoning is essential for \advlongbench, and the size of the
reasoning-induced gain differs sharply by base model. (a) Opus 4.6 with
no reasoning retains non-trivial accuracy on lex-locate baselines
(25--35\%) and on clustered-decoy variants (30--35\%). (b) GPT-5.4 with
no reasoning collapses to $\leq 5\%$ on every variant except the
trivially solvable Unary-sparse baseline. Variants are sorted by Opus 4.6
high-reasoning accuracy (top to bottom).}
\label{fig:reasoning}
\end{figure}

Inference compute in the form of number of reasoning tokens is \textit{critical} for this task. This is true for both the vanilla baseline tasks as well as all the variants. While solving the baseline task itself does not necessarily require a large amount of reasoning, the required amount of reasoning explodes as we increase the difficulty of the search space. In \Cref{fig:reasoning}, we show the effect of reasoning effort required on accuracy for multiple models. All models suffer big declines in accuracy in the low/none reasoning category. 

\begin{figure}[H]
\centering
% Two plots side by side
\begin{subfigure}{0.49\textwidth}
    \centering
    \includegraphics[width=\textwidth]{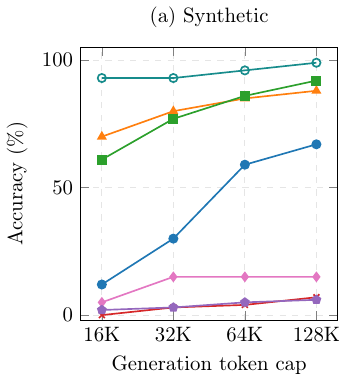}
\end{subfigure}
\hfill
\begin{subfigure}{0.49\textwidth}
    \centering
    \includegraphics[width=\textwidth]{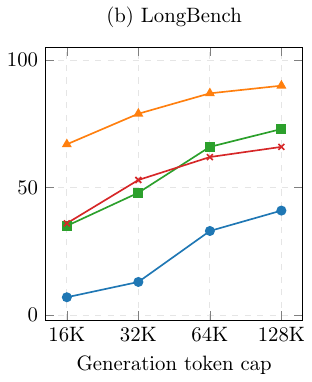}
\end{subfigure}

% Shared legend below
\vspace{0.4em}
\includegraphics[width=0.9\textwidth]{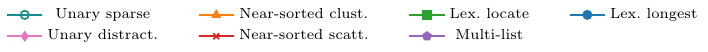}

\caption{GPT-5.4 (high reasoning) accuracy as a function of the generation-token cap. Token caps include reasoning tokens. Tasks vary sharply in how they scale with budget: Unary sparse is saturated at 16K; Lex.\ locate, Near-sorted clustered, and LongBench clustered scale smoothly into the 80--90s; Lex.\ longest and LongBench longest improve dramatically with budget but remain well below baseline locate at 128K; Near-sorted scattered and Multi-list barely move with additional reasoning. Budget alone does not close the decoy gap.}
\label{fig:token_cap}
\end{figure}

Furthermore, in \Cref{fig:token_cap} we show how the accuracy of GPT 5.4 reasoning effort high changes as we increase the inference budget, with clear improvements as the number of output tokens is scaled. However, we note that the accuracy of the baseline task always remains substantially higher than that of the variants. At no point does the latter catch up to the former.

\subsection{Is the model even locating the sequence correctly?}

A plausible failure case for a model is that it is able to locate a sequence correctly but is then unable to fully reproduce it. To test this theory, we compute the accuracy of the first word for the sequence outputted by the model in the cases where the model is wrong. The accuracy is 0\% across all models, suggesting that the model is struggling with locating the correct sequence rather than reproducing it.

\subsection{Opus dominates, OSS models mostly fail}

Opus 4.6 is the strongest model for most task variants (see \Cref{tab:main_results}). While the model is hardly immune to drastic performance drops as we increase the hardness of the problem, it still generally remains better than the other models, achieving the best score in 9/12 task variants. On the other hand, open-source models almost completely fail. While they have respectable scores for the unary predicate baseline task and the $\forall$ quantifier variant, for every other task, they fail to move beyond single-digit accuracy or extremely low double-digit accuracy, often having 0 accuracy.

\subsection{Causes of failures}

The main cause of failure across models is hitting the token budget. Increasing the maximum number of output tokens improves the performance of GPT 5.4, as shown in \Cref{fig:token_cap}. However this is not always true. We took a sample of 5 queries that Opus ran out of tokens for under the 16K budget and gave it a budget of 64K. For some of the queries, it still ran out of budget and for others it generated ~50K--60K tokens before refusing to answer. 

As closed-source models do not reveal their full reasoning trace, we instead turn to the reasoning traces of the frontier OSS models for clues. The traces reveal that models sometimes zone in on the correct location for the target sequence but then suffer from uncertainty and degenerate into repeating the same line again and again. Other times, they reason about various locations in the list and then proceed to output large sections of the list word for word. We have tried without success a few different prompts to push models to avoid outputting large portions of the input list word for word.

\section{Discussion}
In this work, we propose \advlongbench, a new synthetic long-context benchmark that consists of algorithmically simple yet challenging tasks. The basic task provides a context consisting of a list of strings/words and involves searching for certain contiguous subsequences of words in the text. We create several variants of baseline tasks by varying difficulty across several axes: quantifier complexity, adversarial decoys, search space, etc. and show that, despite the relative simplicity of our suite of tasks, frontier LLMs struggle on these tasks. Furthermore, the simplicity of the tasks in \advlongbench  is witnessed by the fact that they do not require LLM-based generations or judges.

We conduct evaluations of several closed-source and open-source frontier LLMs on \advlongbench. The results show that for tasks with unary predicates, the closed-source models (particularly with high reasoning) are reasonably performant on the baseline task as well as the variant with increased quantifier complexity. However, the performance drops to near-zero upon the introduction of scattered adversarial decoys. A similar trend is observed for open-source models.

For the tasks with binary predicates, there is a clear divide between closed-source and open-source models, with the latter having near-zero performance on all variants (including the baseline). Among the closed-source models, Opus 4.6 manages to do the best, with the high reasoning model achieving near-perfect performance on the variant with clustured decoys.

Our investigation of various axes for increasing difficulty of the tasks reveals a number of interesting insights. We show that increasing the arity of predicates is indeed a lever for increasing task difficulty, while increasing quantifier hardness is another means of increasing the difficulty of retrieval (by forcing the model to check a wider range of the context), even when the answer set does not change! Furthermore, our work investigates the use of decoys and shows that their insertion leads to a surprisingly drastic drop in performance across all models.

Our work leaves open the possibility of future avenues of investigation. \advlongbench consists of tasks on unary (prefix/suffix/containment) and binary (lexicographic) predicates. However, one can experiment with more complex unary/binary predicates or consider $k$-ary predicates for $k > 2$, which can stress test long-context capabilities with more complex computation.

A limitation of our work is that in the quantifier complexity investigations, we have not constructed tasks where there is a longer list of sequences satisfying the given constraints. An exciting future direction would be to stress test long-context capabilities by increasing the size of complete set of satisfying sequences. Furthermore, our work could benefit from a broader analysis of decoy variants and their arrangement in the context.

Our hope is that an understanding of models' limitations on the kinds of tasks in \advlongbench can yield insights into improving long-context capabilities of future models.

\section{Acknowledgements}

Thanks to Boris Ginsburg, Navya Ilineni, and Krishna Puvvada for useful discussions and feedback.

\bibliographystyle{plainnat}
\bibliography{references}

%%%%%%%%%%%%%%%%%%%%%%%%%%%%%%%%%%%%%%%%%%%%%%%%%%%%%%%%%%%%

\newpage

\appendix

\section{Dataset Construction}
\label{app:tasks}

This appendix gives full procedures for generating each task variant in
\advlongbench. Throughout, we write $\prec$ for the standard lexicographic
order on strings, $\mathcal{W}$ for the global pool of available distinct
words, $k$ for target sequence length, and $N$ for total list length. All
sampling is uniform unless otherwise stated.

\subsection{Synthetic word sampling}

For synthetic tasks we sample distinct random strings of fixed length $L$ over
the lowercase alphabet. We sample $N$ integers without replacement from
$\{0, \dots, 26^L - 1\}$ and map each to a unique base-$26$ string. We use
$L = 8$ for lexicographic tasks and $L = 12$ for unary-predicate tasks.

\begin{algorithm}[H]
\caption{\textsc{SampleSyntheticWords}}
\label{alg:sample-words}
\begin{algorithmic}[1]
\Require word count $N$, word length $L$
\Ensure list $W$ of $N$ distinct length-$L$ strings over $\{a, \dots, z\}$
\State $S \gets$ sample $N$ integers without replacement from $\{0, \dots, 26^L - 1\}$
\State $W \gets [\textsc{Base26}(s, L) : s \in S]$ \Comment{base-$26$ encoding to letters}
\State shuffle $W$
\State \Return $W$
\end{algorithmic}
\end{algorithm}

\subsection{Lexicographic tasks (synthetic)}
\label{app:lex-synthetic}

\subsubsection{Plant-block construction}

The target lexicographic run is inserted as a length-$(k{+}2)$ block whose
middle $k$ positions form a sorted run, while the two flanking positions
deliberately violate lex order so the run cannot be extended.

\begin{algorithm}[H]
\caption{\textsc{ConstructPlantBlock}}
\label{alg:plant-block}
\begin{algorithmic}[1]
\Require word pool $\mathcal{W}$, target length $k$
\Ensure block $B$ of length $k+2$ such that $B[1..k]$ is sorted and
        the run cannot be extended to either flank
\State sample $u_0, u_1, \dots, u_{k+1}$ from $\mathcal{W}$ without replacement
\State sort so $u_0 \prec u_1 \prec \cdots \prec u_{k+1}$
\State $B \gets [\,u_{k+1},\; u_1, u_2, \dots, u_k,\; u_0\,]$
\State \Return $B$
\end{algorithmic}
\end{algorithm}

\noindent
By construction $B[0] = u_{k+1} \succ B[1]$ and $B[k] \succ B[k+1] = u_0$, so
neither flank extends the run, while $B[1] \prec B[2] \prec \cdots \prec B[k]$
gives the unique valid lex run of length $k$ within $B$.

\subsubsection{Baseline \textsc{Locate} instance}

We then plant the block at a random position in a shuffled list of synthetic
words and iteratively destroy any unintended lex run of length $\geq k$ that
arises elsewhere by chance.

\begin{algorithm}[H]
\caption{\textsc{GenerateSyntheticLexInstance}}
\label{alg:synth-lex}
\begin{algorithmic}[1]
\Require list size $N$, target length $k$
\Ensure word list $W$ with a unique lex run of length $k$ at indices $\mathcal{T}$
\State $W \gets \textsc{SampleSyntheticWords}(N, L{=}8)$
\State $B \gets \textsc{ConstructPlantBlock}(\mathcal{W}, k)$
\State sample insertion index $i \in \{0, \dots, N - (k+2)\}$
\State $W[i : i + k + 2] \gets B$
\State $\mathcal{T} \gets \{i+1, \dots, i+k\}$
\While{$\exists$ maximal lex run $[a, b] \subseteq [0, N{-}1]$
       with $b - a + 1 \geq k$ and $\{a, \dots, b\} \neq \mathcal{T}$}
  \State pick any $j \in [a+1, b]$
  \State sample $j' > j$ such that $W[j'] \prec W[j-1]$
  \State swap $W[j]$ and $W[j']$ \Comment{breaks the run at position $j$}
\EndWhile
\State \Return $(W, \mathcal{T})$
\end{algorithmic}
\end{algorithm}

\subsubsection{\textsc{Longest} sequence variant}

The \textsc{Longest sequence retrieval} task uses Algorithm~\ref{alg:synth-lex} unchanged: the
planted run of length $k$ is, by construction, the unique longest lex run in
$W$. Only the prompt differs --- the model is asked for the longest valid
run rather than a run of given length.

\subsubsection{Near-sorted decoys}

A near-sorted decoy is a length-$(k{+}2)$ block built like a plant block, but
with one interior pair swapped. The block visually resembles a planted run
and the surrounding flanks still prevent extension; the only failure is an
internal inversion.

\begin{algorithm}[H]
\caption{\textsc{ConstructNearSortedDecoy}}
\label{alg:near-sorted}
\begin{algorithmic}[1]
\Require word pool $\mathcal{W}$, target length $k$
\Ensure length-$(k{+}2)$ near-sorted decoy block $B$
\State sample and sort $u_0 \prec u_1 \prec \cdots \prec u_{k+1}$ from $\mathcal{W}$
\State $B \gets [\,u_{k+1},\; u_1, \dots, u_k,\; u_0\,]$
\State sample distinct $p, q \in \{1, \dots, k\}$
\State swap $B[p]$ and $B[q]$
\State \Return $B$
\end{algorithmic}
\end{algorithm}

\noindent
The \emph{scattered} variant inserts $D = 8$ decoys at uniformly random
non-overlapping positions; the \emph{clustered} variant places them within a
small radius $\rho$ of the planted target. After insertion, unintended lex
runs that arise outside the target and the decoys are destroyed using the
fixup loop of Algorithm~\ref{alg:synth-lex}.

\subsubsection{Multi-list variant}

The list of $N$ words is partitioned into $M = 20$ contiguous sublists. The
target run is planted within one sublist as in
Algorithm~\ref{alg:synth-lex}. At every adjacent boundary $(L_t, L_{t+1})$ we
additionally plant a \emph{cross-boundary} length-$k$ sorted run: the suffix
of $L_t$ together with the prefix of $L_{t+1}$ would form a valid run if
the lists were concatenated. The model is instructed to return a run
contained \emph{within a single sublist}, so cross-boundary runs are decoys.

\begin{algorithm}[H]
\caption{\textsc{GenerateMultiListInstance}}
\label{alg:multi-list}
\begin{algorithmic}[1]
\Require total length $N$, target length $k$, number of sublists $M$,
         boundary split $(k_1, k_2)$ with $k_1 + k_2 = k$
\Ensure sublists $L_1, \dots, L_M$ with a single within-sublist target run
\State $W \gets \textsc{SampleSyntheticWords}(N, L{=}8)$
\State partition $W$ into contiguous sublists $L_1, \dots, L_M$
\State pick target sublist $m^* \in \{1, \dots, M\}$ at random
\State plant the target block in $L_{m^*}$ via Alg.~\ref{alg:plant-block}
\For{$t = 1, \dots, M-1$}
  \State sample sorted $u_0 \prec \cdots \prec u_{k-1}$ from $\mathcal{W}$
  \State write $u_0, \dots, u_{k_1 - 1}$ into the last $k_1$ positions of $L_t$
  \State write $u_{k_1}, \dots, u_{k-1}$ into the first $k_2$ positions of $L_{t+1}$
  \State adjust the position immediately preceding (resp.\ following) the
         boundary so the run does not extend further within $L_t$ (resp.\ $L_{t+1}$)
\EndFor
\State destroy any unintended within-sublist lex run of length $\geq k$
       outside the target via the fixup loop of Alg.~\ref{alg:synth-lex}
\State \Return $L_1, \dots, L_M$, target sublist $m^*$, target indices
\end{algorithmic}
\end{algorithm}

\subsection{Unary-predicate tasks}
\label{app:unary}

Unary tasks use $12$-letter synthetic words and predicates of the form
$(p, s)$ where $p \in \{\textsc{prefix}, \textsc{suffix}, \textsc{contains}\}$
and $s$ is a length-$m$ substring (we use $m = 2$). For target length $k$
we sample one predicate per target position. The target window is the
unique length-$k$ contiguous window in which position $j$ satisfies
predicate $j$ for all $j$. The distractor variant additionally plants
predicate matches at a fraction $\alpha$ of non-target positions, subject
to two constraints: (i) no non-target word satisfies more than one
predicate, and (ii) no length-$k$ contiguous window outside the target
satisfies all $k$ predicates in order.

\begin{algorithm}[H]
\caption{\textsc{GenerateUnaryPredicateInstance}}
\label{alg:unary}
\begin{algorithmic}[1]
\Require list size $N$, target length $k$, predicate string length $m$,
         distractor fraction $\alpha \geq 0$
\Ensure list $W$, predicates $\{(p_j, s_j)\}_{j=1}^{k}$, target indices $\mathcal{T}$
\State $W \gets \textsc{SampleSyntheticWords}(N, L{=}12)$
\For{$j = 1, \dots, k$}
  \State $p_j \sim \text{Unif}\{\textsc{prefix}, \textsc{suffix}, \textsc{contains}\}$
  \State $s_j \sim \text{Unif}(\{a, \dots, z\}^m)$
\EndFor
\State sample insertion index $i \in \{0, \dots, N - k\}$;
       $\mathcal{T} \gets \{i, \dots, i + k - 1\}$
\For{$j = 1, \dots, k$}
  \State edit $W[i + j - 1]$ minimally so it satisfies $(p_j, s_j)$
\EndFor
\If{$\alpha = 0$} \Comment{baseline: no non-target match}
  \State scrub any non-target word that incidentally matches some $(p_j, s_j)$
\Else \Comment{distractor variant}
  \For{each predicate $(p_j, s_j)$}
    \State $\textsc{count} \gets \lfloor \alpha \cdot (N - k) \rfloor$
    \While{fewer than $\textsc{count}$ distractors planted}
      \State sample $r \in \{0, \dots, N-1\} \setminus \mathcal{T}$
      \State $W' \gets W$ with $W'[r]$ edited to satisfy $(p_j, s_j)$
      \If{$W'[r]$ now satisfies two distinct predicates} \textbf{continue} \EndIf
      \If{$W'$ contains any length-$k$ window outside $\mathcal{T}$ matching all
          $(p_1, s_1), \dots, (p_k, s_k)$ in order} \textbf{continue} \EndIf
      \State $W \gets W'$
    \EndWhile
  \EndFor
\EndIf
\State \Return $W, \{(p_j, s_j)\}_{j=1}^k, \mathcal{T}$
\end{algorithmic}
\end{algorithm}

\subsection{LongBench-based lexicographic tasks}
\label{app:longbench}

The natural-text variants draw word lists from documents in LongBench~v2,
filtered and post-processed as described below.

\subsubsection{Document source distribution}

We sample documents across the LongBench~v2 task categories shown in
\Cref{tab:longbench-sources}. Counts reflect documents retained after
the filtering stage of \S\ref{app:lb-filter}.

\begin{table}[h]
\centering
\caption{LongBench~v2 source-task breakdown of documents used in
\advlongbench (post-filtering).}
\label{tab:longbench-sources}
\small
\begin{tabular}{lr}
\toprule
LongBench~v2 source task & Count \\
\midrule
Single-Document QA / Literary       & 19 \\
Multi-Document QA / Multi-news      & 13 \\
Single-Document QA / Academic       & 12 \\
Single-Document QA / Legal          & 11 \\
Single-Document QA / Governmental   & 9  \\
Single-Document QA / Financial      & 8  \\
Multi-Document QA / Governmental    & 8  \\
Multi-Document QA / Academic        & 7  \\
Multi-Document QA / Legal           & 6  \\
Long In-context Learning / User-guide QA & 3 \\
Single-Document QA / Detective      & 3  \\
Multi-Document QA / Financial       & 1  \\
\midrule
Total                               & 100 \\
\bottomrule
\end{tabular}
\end{table}

\subsubsection{Document filtering}
\label{app:lb-filter}

Documents are tokenized into a flat word list and then filtered to remove
artifacts that would let the task be solved without genuine long-context
search (e.g., a near-sorted run of length 30 already present in the source,
or a run dominated by a single repeated character or word).

\begin{algorithm}[H]
\caption{\textsc{FilterLongBenchDocument}}
\label{alg:filter}
\begin{algorithmic}[1]
\Require document $d$, max longest-run threshold $k_{\max}$ (default $10$),
         min run length to audit $k_{\min}$ (default $5$)
\Ensure boolean: keep or discard $d$
\If{$d$ contains HTML tags} \Return discard \EndIf
\State $W \gets$ tokenize($d$)
\State $\mathcal{R} \gets$ all maximal lex runs in $W$ of length $\geq k_{\min}$
\State $R^* \gets \arg\max_{R \in \mathcal{R}} |R|$
\If{$|R^*| > k_{\max}$} \Return discard \EndIf
\For{each $R \in \mathcal{R}$}
  \State $C \gets$ concatenation of all characters of words in $R$
  \If{$\max_c \mathrm{count}(c, C) \,/\, |C| \geq 0.3$} \Return discard \EndIf
  \If{$\max_w \mathrm{count}(w, R) \,/\, |R| \geq 0.2$} \Return discard \EndIf
\EndFor
\If{$R^*$ contains duplicate words} \Return discard \EndIf
\If{any word in $R^*$ contains an apostrophe} \Return discard \EndIf
\State \Return keep
\end{algorithmic}
\end{algorithm}

\subsubsection{Near-sorted decoy insertion}

For LongBench-based decoy variants, decoys are constructed using words drawn
from the document itself, so that the decoys remain
distributionally consistent with their natural-text surroundings and cannot
be detected through superficial cues (e.g., out-of-vocabulary strings).

\begin{algorithm}[H]
\caption{\textsc{InsertLongBenchNearSortedDecoy}}
\label{alg:lb-decoy}
\begin{algorithmic}[1]
\Require word list $W$ from a filtered LongBench document, target length $k$
\Ensure $W$ with one near-sorted decoy of length $k$ planted at a natural boundary
\State sample boundary index $b$ such that $W[b{-}1] \succ W[b]$
\State $\mathcal{V} \gets$ unique words in $W$
\State sample $w_{\text{lo}} \in \mathcal{V}$ with $w_{\text{lo}} \prec W[b{-}1]$
\State sample $w_{\text{hi}} \in \mathcal{V}$ with $w_{\text{hi}} \succ W[b]$
\If{$w_{\text{hi}} \prec w_{\text{lo}}$} swap $w_{\text{lo}} \leftrightarrow w_{\text{hi}}$ \EndIf
\State sample $u_1, \dots, u_{k-2} \in \mathcal{V}$ with
       $w_{\text{lo}} \preceq u_j \preceq w_{\text{hi}}$
\State sort so $u_1 \preceq u_2 \preceq \cdots \preceq u_{k-2}$
\State sample distinct $p, q \in \{1, \dots, k-2\}$ and swap $u_p \leftrightarrow u_q$
\State $D \gets [\, w_{\text{lo}},\; u_1, \dots, u_{k-2},\; w_{\text{hi}} \,]$
\State insert $D$ into $W$ between positions $b-1$ and $b$
\State \Return $W$
\end{algorithmic}
\end{algorithm}

\noindent
The boundary condition $W[b{-}1] \succ W[b]$ ensures that the surrounding
context cannot extend the planted near-sorted run past either end. The
\emph{scattered} variant repeats Algorithm~\ref{alg:lb-decoy} for $D = 8$
boundaries chosen at random; the \emph{clustered} variant restricts the
boundaries to a window around the target run.

\section{Prompts}

\begin{figure}[H]
  \centering
  \includegraphics[width=\linewidth]{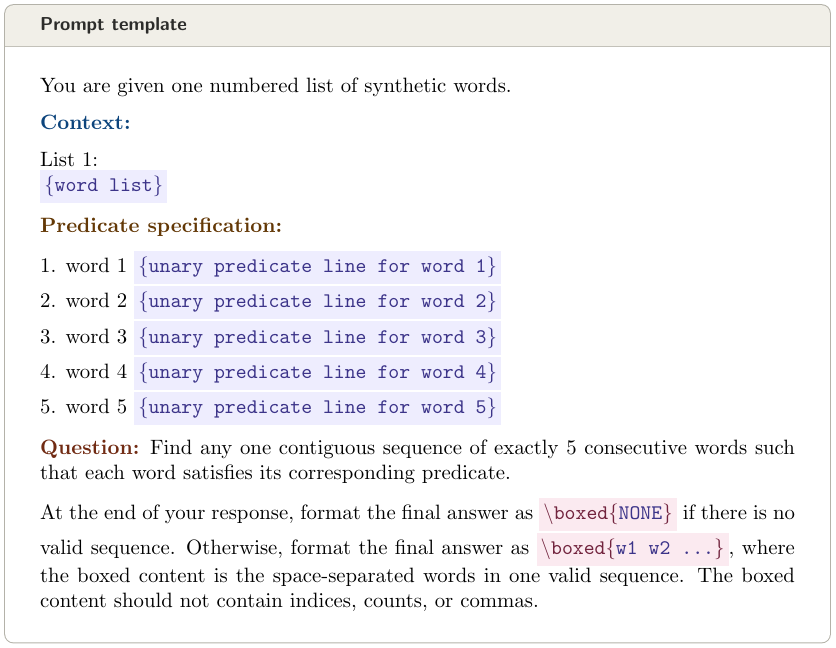}
  \caption{Prompt template for unary predicate where we want to extract any satisfying sequence.}
  \label{fig:unary_one_prompt}
\end{figure}

\begin{figure}[H]
  \centering
  \includegraphics[width=\linewidth]{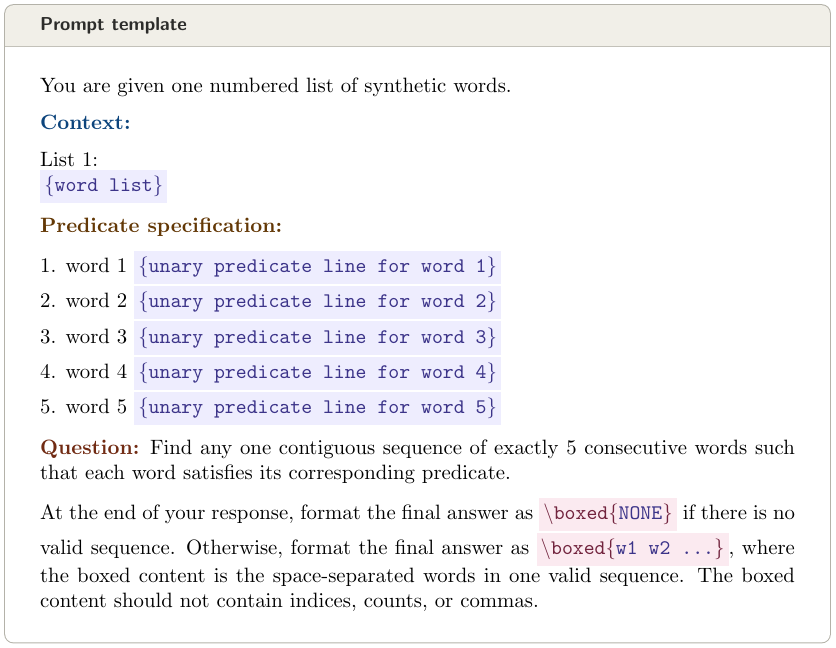}
  \caption{Prompt template for unary predicate where we want to extract all satisfying sequences.}
  \label{fig:unary_all_prompts}
\end{figure}

\begin{figure}[H]
  \centering
  \includegraphics[width=\linewidth]{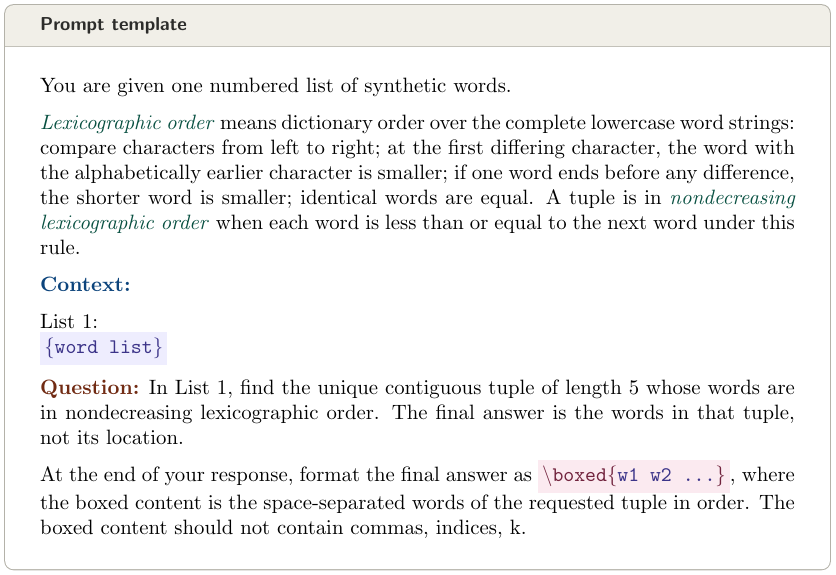}
  \caption{Prompt template for lexicographic predicate where we want to extract any satisfying sequence.}
  \label{fig:lex_one_prompt}
\end{figure}

\begin{figure}[H]
  \centering
  \includegraphics[width=\linewidth]{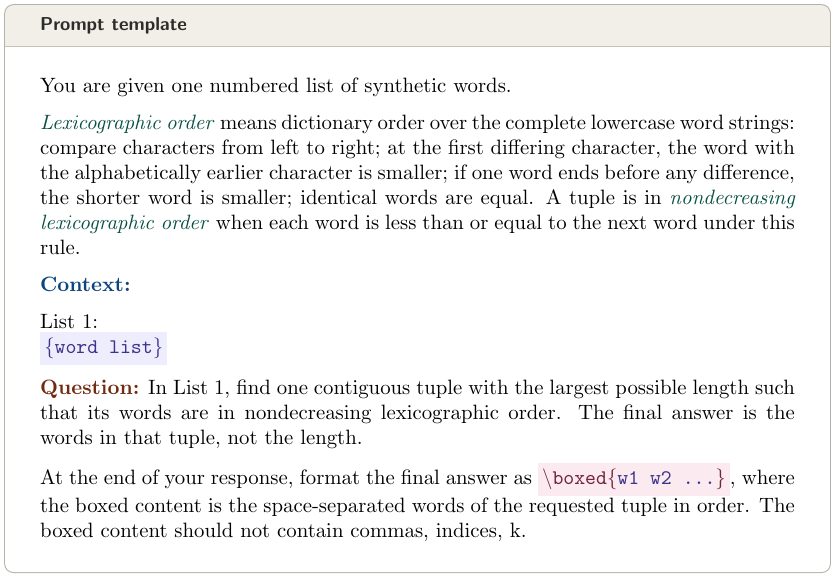}
  \caption{Prompt template for lexicographic predicate where we want to extract the longest satisfying sequence.}
  \label{fig:lex_longest_prompt}
\end{figure}

\begin{figure}[H]
  \centering
  \includegraphics[width=\linewidth]{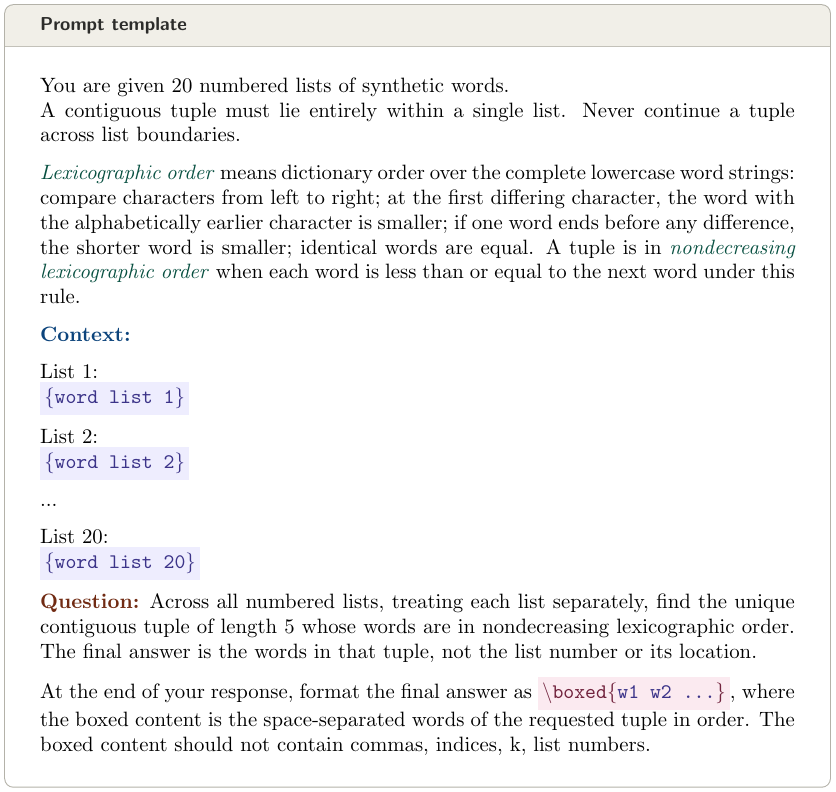}
  \caption{Prompt template for lexicographic predicate where we want to extract any satisfying sequence from multiple lists.}
  \label{fig:lex-prompt}
\end{figure}

\section{Inference Setup}

For all models, we set max output tokens to 16K (16384) unless otherwise specified.

\paragraph{GPT 5.4} We call the API endpoint with reasoning effort high and none. For both queries we requested 128K max output tokens. Everywhere where we report results for 16K output tokens, we use the responses from these queries and consider responses > 16K tokens as incorrect.

\paragraph{Opus 4.6} We call Opus 4.6 with max output tokens 16K and a thinking budget of 12K for high and 0 for none.

\paragraph{Gemini 3.1 Pro Preview} We call it with max output tokens 16K and a thinking budget of 12K for high and 1K for low (our API endpoint did not allow us to go lower)

\paragraph{MiniMax-2.7, Qwen3.5 397B, GLM-5.1} We ran it with temperature 1.0, top-p 1.0, and max output tokens 16K. We ran MiniMax on A100, Qwen3.5 on B200, and GLM-5.1 on H100 GPUs.

\section{Additional Figures}

%%%%%%%%%%%%%%%%%%%%%%%%%%%%%%%%%%%%%%%%%%%%%%%%%%%%%%%%%%%%

\clearpage
\section*{NeurIPS Paper Checklist}

%%% BEGIN INSTRUCTIONS %%%
The checklist is designed to encourage best practices for responsible machine learning research, addressing issues of reproducibility, transparency, research ethics, and societal impact. Do not remove the checklist: {\bf The papers not including the checklist will be desk rejected.} The checklist should follow the references and follow the (optional) supplemental material.  The checklist does NOT count towards the page
limit. 

Please read the checklist guidelines carefully for information on how to answer these questions. For each question in the checklist:
\begin{itemize}
    \item You should answer \answerYes{}, \answerNo{}, or \answerNA{}.
    \item \answerNA{} means either that the question is Not Applicable for that particular paper or the relevant information is Not Available.
    \item Please provide a short (1--2 sentence) justification right after your answer (even for \answerNA). 
   % \item {\bf The papers not including the checklist will be desk rejected.}
\end{itemize}

{\bf The checklist answers are an integral part of your paper submission.} They are visible to the reviewers, area chairs, senior area chairs, and ethics reviewers. You will also be asked to include it (after eventual revisions) with the final version of your paper, and its final version will be published with the paper.

The reviewers of your paper will be asked to use the checklist as one of the factors in their evaluation. While \answerYes{} is generally preferable to \answerNo{}, it is perfectly acceptable to answer \answerNo{} provided a proper justification is given (e.g., error bars are not reported because it would be too computationally expensive'' or ``we were unable to find the license for the dataset we used''). In general, answering \answerNo{} or \answerNA{} is not grounds for rejection. While the questions are phrased in a binary way, we acknowledge that the true answer is often more nuanced, so please just use your best judgment and write a justification to elaborate. All supporting evidence can appear either in the main paper or the supplemental material, provided in appendix. If you answer \answerYes{} to a question, in the justification please point to the section(s) where related material for the question can be found.

IMPORTANT, please:
\begin{itemize}
    \item {\bf Delete this instruction block, but keep the section heading ``NeurIPS Paper Checklist"},
    \item  {\bf Keep the checklist subsection headings, questions/answers and guidelines below.}
    \item {\bf Do not modify the questions and only use the provided macros for your answers}.
\end{itemize}

%%% END INSTRUCTIONS %%%

\begin{enumerate}

\item {\bf Claims}
    \item[] Question: Do the main claims made in the abstract and introduction accurately reflect the paper's contributions and scope?
    \item[] Answer: \answerYes{} % Replace by \answerYes{}, \answerNo{}, or \answerNA{}.
    \item[] Justification: We claim that we define and evaluate various axes of difficulty for long context tasks and we indeed do so

\item {\bf Limitations}
    \item[] Question: Does the paper discuss the limitations of the work performed by the authors?
    \item[] Answer: \answerYes{} % Replace by \answerYes{}, \answerNo{}, or \answerNA{}.
    \item[] Justification: We discuss limitations in the Discussion section

\item {\bf Theory assumptions and proofs}
    \item[] Question: For each theoretical result, does the paper provide the full set of assumptions and a complete (and correct) proof?
    \item[] Answer: \answerNA{} % Replace by \answerYes{}, \answerNo{}, or \answerNA{}.
    \item[] Justification: We do not have any theoretical results.

    \item {\bf Experimental result reproducibility}
    \item[] Question: Does the paper fully disclose all the information needed to reproduce the main experimental results of the paper to the extent that it affects the main claims and/or conclusions of the paper (regardless of whether the code and data are provided or not)?
    \item[] Answer: \answerYes{} % Replace by \answerYes{}, \answerNo{}, or \answerNA{}.
    \item[] Justification: We describe in detail the construction of the dataset as well as the prompts used for querying the LLMs, and the max token lengths requested.

\item {\bf Open access to data and code}
    \item[] Question: Does the paper provide open access to the data and code, with sufficient instructions to faithfully reproduce the main experimental results, as described in supplemental material?
    \item[] Answer: \answerYes{} % Replace by \answerYes{}, \answerNo{}, or \answerNA{}.
    \item[] Justification: We provide our full prompts dataset as well as the code to generate, validate, and evaluate tasks for our benchmark.

\item {\bf Experimental setting/details}
    \item[] Question: Does the paper specify all the training and test details (e.g., data splits, hyperparameters, how they were chosen, type of optimizer) necessary to understand the results?
    \item[] Answer: \answerYes{} % Replace by \answerYes{}, \answerNo{}, or \answerNA{}.
    \item[] Justification: We do not train models but we provide the details of how we queired the LLMs.

\item {\bf Experiment statistical significance}
    \item[] Question: Does the paper report error bars suitably and correctly defined or other appropriate information about the statistical significance of the experiments?
    \item[] Answer: \answerNo{} % Replace by \answerYes{}, \answerNo{}, or \answerNA{}.
    \item[] Justification: The effects we analyze are very large and far outside the error bars.

\item {\bf Experiments compute resources}
    \item[] Question: For each experiment, does the paper provide sufficient information on the computer resources (type of compute workers, memory, time of execution) needed to reproduce the experiments?
    \item[] Answer: \answerYes{} % Replace by \answerYes{}, \answerNo{}, or \answerNA{}.
    \item[] Justification: We query closed LLMs through an API. We ran open LLMs on our own GPUs and describe our setup.
    
\item {\bf Code of ethics}
    \item[] Question: Does the research conducted in the paper conform, in every respect, with the NeurIPS Code of Ethics \url{https://neurips.cc/public/EthicsGuidelines}?
    \item[] Answer: \answerYes{} % Replace by \answerYes{}, \answerNo{}, or \answerNA{}.
    \item[] Justification: Our research does not involve human participants. The datasets we generate are synthetic or reuse an existing published dataset (LongBench v2) which is Apache 2.0 licensed. They do not have any privacy or consent or copyright issues.

\item {\bf Broader impacts}
    \item[] Question: Does the paper discuss both potential positive societal impacts and negative societal impacts of the work performed?
    \item[] Answer: \answerNo{} % Replace by \answerYes{}, \answerNo{}, or \answerNA{}.
    \item[] Justification: The paper is about systematic evaluation of long context ability of LLMs on synthetic tasks and thus does not have societal impact.
    
\item {\bf Safeguards}
    \item[] Question: Does the paper describe safeguards that have been put in place for responsible release of data or models that have a high risk for misuse (e.g., pre-trained language models, image generators, or scraped datasets)?
    \item[] Answer: \answerNA{} % Replace by \answerYes{}, \answerNo{}, or \answerNA{}.
    \item[] Justification: The dataset is synthetically generated and does not involve any sensitive or dangerous information.

\item {\bf Licenses for existing assets}
    \item[] Question: Are the creators or original owners of assets (e.g., code, data, models), used in the paper, properly credited and are the license and terms of use explicitly mentioned and properly respected?
    \item[] Answer: \answerYes{} % Replace by \answerYes{}, \answerNo{}, or \answerNA{}.
    \item[] Justification: We use part of the LongBench v2 dataset and credit them

\item {\bf New assets}
    \item[] Question: Are new assets introduced in the paper well documented and is the documentation provided alongside the assets?
    \item[] Answer: \answerYes{} % Replace by \answerYes{}, \answerNo{}, or \answerNA{}.
    \item[] Justification: We provide a readme along with our dataset that describes it. We also have a readme for the code.

\item {\bf Crowdsourcing and research with human subjects}
    \item[] Question: For crowdsourcing experiments and research with human subjects, does the paper include the full text of instructions given to participants and screenshots, if applicable, as well as details about compensation (if any)? 
    \item[] Answer: \answerNA{} % Replace by \answerYes{}, \answerNo{}, or \answerNA{}.
    \item[] Justification: We did not crowdsource or research with human subjects.

\item {\bf Institutional review board (IRB) approvals or equivalent for research with human subjects}
    \item[] Question: Does the paper describe potential risks incurred by study participants, whether such risks were disclosed to the subjects, and whether Institutional Review Board (IRB) approvals (or an equivalent approval/review based on the requirements of your country or institution) were obtained?
    \item[] Answer: \answerNA{} % Replace by \answerYes{}, \answerNo{}, or \answerNA{}.
    \item[] Justification: We did not work with human or living subjects.

\item {\bf Declaration of LLM usage}
    \item[] Question: Does the paper describe the usage of LLMs if it is an important, original, or non-standard component of the core methods in this research? Note that if the LLM is used only for writing, editing, or formatting purposes and does \emph{not} impact the core methodology, scientific rigor, or originality of the research, declaration is not required.
    %this research? 
    \item[] Answer: \answerNA{} % Replace by \answerYes{}, \answerNo{}, or \answerNA{}.
    \item[] Justification: LLMs were only used for some code writing, and figure generation. They were not involved in figuring out the core methodology of the paper or the setup of experiments.

\end{enumerate}

\end{document}